%% file: main.tex
\documentclass[final]{cvpr}

\usepackage{times}
\usepackage{epsfig}
\usepackage{graphicx}
\usepackage{amsmath}
\usepackage{amssymb}
\usepackage{verbatim}

\usepackage{pifont} 
\usepackage{multirow} 
\usepackage{booktabs} 

\newcommand{\webpage}{\url{https://bit.ly/3dg90fV}}
\newcommand{\githubpage}{\url{https://bit.ly/3t66bnU}}

\usepackage[pagebackref=true,breaklinks=true,colorlinks,bookmarks=false]{hyperref}

\def\cvprPaperID{222} 
\def\confYear{CVPR 2021}

\begin{document}

\title{Stochastic Image-to-Video Synthesis using cINNs}

\author{
Michael Dorkenwald$^1$ \quad Timo Milbich$^1$ \quad Andreas Blattmann$^1$ \quad Robin Rombach$^1$ \\
Konstantinos G. Derpanis$^{2,3,4}$\thanks{Indicates equal supervision.} \quad Bj{\"o}rn Ommer$^1$\footnote[1]{}\\
\normalsize{$^1$IWR/HCI, Heidelberg University, Germany \quad $^2$Department of Computer Science, Ryerson University, Canada} \\
\normalsize{$^3$Vector Institute for AI, Canada \quad $^4$Samsung AI Centre Toronto, Canada}
}

\maketitle

\begin{abstract}
Video understanding calls for a model to learn the characteristic interplay between static scene content and its dynamics: Given an image, the model must be able to predict a future progression of the portrayed scene and, conversely, a video should be explained in terms of its static image content and all the remaining characteristics not present in the initial frame. This naturally suggests a bijective mapping between the video domain and the static content as well as residual information. In contrast to common stochastic image-to-video synthesis, such a model does not merely generate arbitrary videos progressing the initial image. Given this image, it rather provides a one-to-one mapping between the residual vectors and the video with stochastic outcomes when sampling. The approach is naturally implemented using a conditional invertible neural network (cINN) that can explain videos by independently modelling static and other video characteristics, thus laying the basis for controlled video synthesis. Experiments on four diverse video datasets demonstrate the effectiveness of our approach in terms of both the quality and diversity of the synthesized results. Our project page is available at \small{\webpage}.
\end{abstract}

\vspace{-0.3cm}
\section{Introduction}
\input{chapters/introduction}

\section{Related Work}
\input{chapters/related_work}

\section{Method}
\input{chapters/method}

\section{Experiments}
\input{chapters/results}

\section{Conclusion}
\input{chapters/conclusion}

{\small
\bibliographystyle{ieee_fullname}
\bibliography{short-definitions, egbib}
}

\clearpage
\newpage

{\noindent \Huge \textbf{Supplemental}}

\appendix

\input{supplementary}

\end{document}

%% file: chapters/introduction.tex
Anticipating and predicting what happens next are key features of human intelligence that allow us to understand and deal with the ever-changing environment that governs our everyday life \cite{pcb_ricarda}. Consequently, the ability to foresee and hallucinate the future progression of a scene is a cornerstone of artificial visual understanding with applications including autonomous driving~\cite{martin2019drive,masood2018detecting,gebert2019end}, medical treatment~\cite{uBAM_Biagio, magnification_dorkenwald, LSTM_biagio}, and robotic planning~\cite{bair, finn2017deep, byravan2017se3posenets}. 
\begin{figure}
\begin{center}
\includegraphics[width=0.49\textwidth]{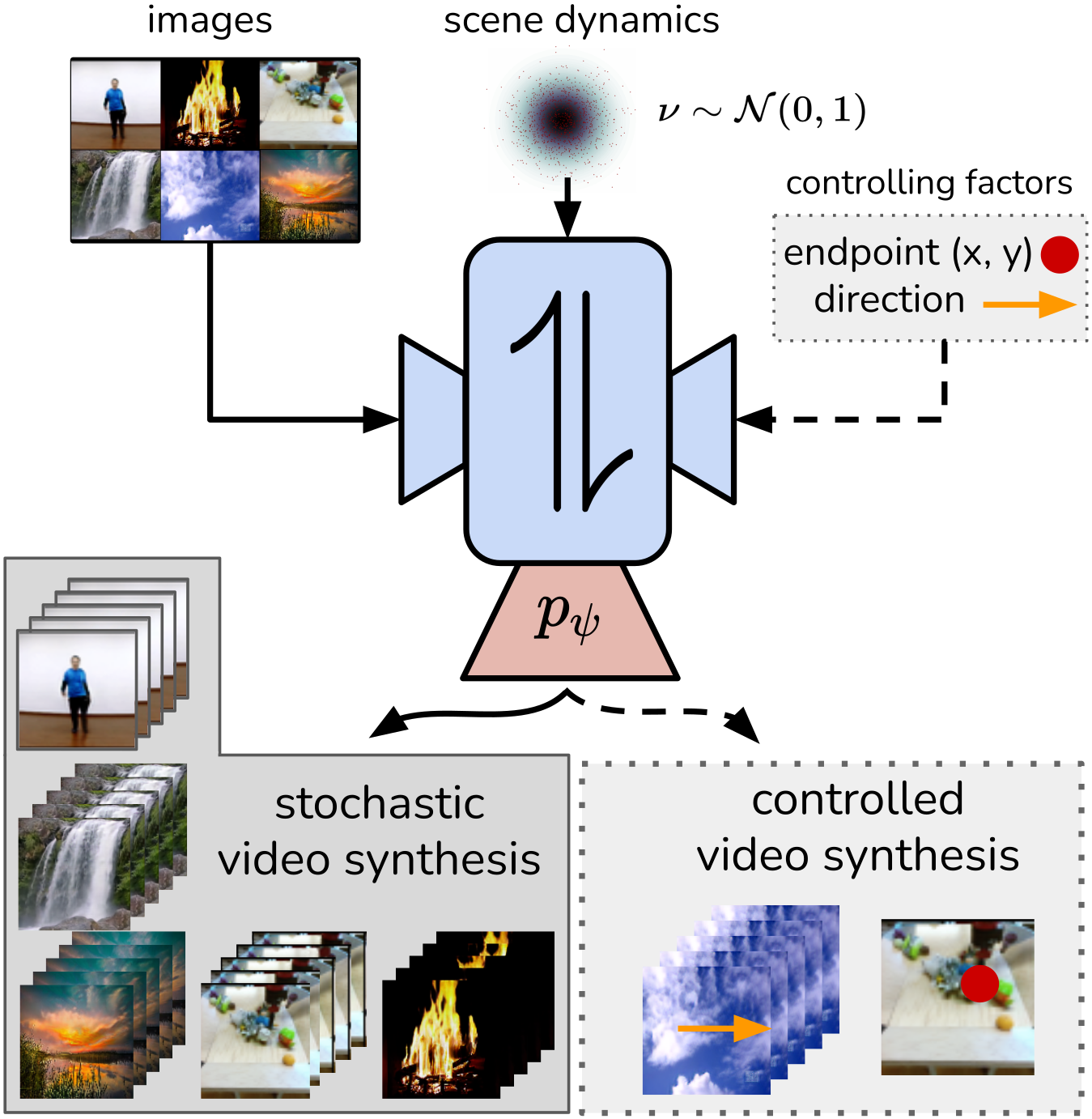}
\caption{Our approach establishes a bijective mapping between the image and the video domain by introducing a residual representation $\nu$ describing the latent scene dynamics. This allows us not only to synthesize diverse videos but also to extend our approach to gain control over the video synthesis task. 
}\label{fig:first_page}
\vspace{-0.8cm}
\end{center}
\end{figure}

Predicting and synthesizing plausible future progressions from a given image requires a deep understanding of how scenes and objects within video are depicted, interplay with each other, and evolve over time. While an image provides information about the observed scene content, such as object appearance and shape, the challenge is to understand the missing information constituting potential futures, such as the scene dynamics setting the scene in motion. Due to the ambiguity and complexity of capturing this information, many works~\cite{SAVP, franceschi2020stochastic, Castrejon_2019_ICCV, weissenborn2020scaling} directly focus on predicting likely video continuations, often resorting to simplifying assumptions (e.g., dynamics modelled by optical flow \cite{Animating_landscapes, reda2018sdcnet}) and side information (e.g., semantic keypoints \cite{keyin_kosta, walker2017pose,unsup_partbased,hbugen, behavior_driven}). However, truly understanding the synthesis problem not only requires to infer such image continuations but, conversely, also demands when observing a video sequence to describe and represent the instantiated scene dynamics animating its initial frame.

Consequently, the image-to-video synthesis task should be modelled as a translation between the image and video domains, ideally by an invertible mapping between them. 
Since the content information describing an image only accounts for a small fraction of the video information, in particular missing the temporal dimension, learning an invertible mapping requires a dedicated residual representation that captures all missing information. Once learned, given an initial image and an instantiation of the latent residual, we can combine them to synthesize the corresponding future video sequence. 

In this paper, we frame image-to-video synthesis as an invertible domain transfer problem and implement it using a conditional invertible neural network (cINN) illustrated in Fig.~\ref{fig:first_page}. To account for the domain gap between images and videos, we introduce a dedicated probabilistic residual representation. The bijective nature of our mapping ensures that only information complementary to that in the initial image is captured. 
Using a probabilistic formulation,
the residual representation allows to sample and thus synthesize novel future progressions in video with the same start frame. To reduce the complexity of the learning task, we train a separate conditional variational encoder-decoder architecture to compute a compact, information preserving representation for the video domain.
%
Moreover, our specific framing of learning the residual representation allows to easily incorporating extra conditioning information to exercise control over the image-to-video synthesis process.

Our contributions can be summarized as follows:
\begin{itemize}
    \item We frame image-to-video synthesis as an invertible domain transfer problem and learn a dedicated residual representation to capture the domain gap.
    \item Our framework naturally extends to incorporate explicit conditioning factors for exercising control over the synthesis process.
    \item Extensive evaluations on four diverse video datasets, ranging from structured human motion synthesis to subtle dynamic textures, show strong results demonstrating the effectiveness of our approach.
\end{itemize}

%% file: chapters/related_work.tex
\noindent{\bf Video synthesis.} Video synthesis involves a wide range of tasks including video-to-video translation~\cite{vid2vid}, image animation \cite{Siarohin_2019_CVPR,first_order_2019}, frame interpolation \cite{Niklaus_2018_CVPR,Bao_2019_CVPR}, and video prediction. The latter can be divided into unconditional~\cite{tulyakov2017mocogan, DVDGAN} and conditional video generation (the focus of our work). Conditional video generation can be described as finding a future progression given a set of context frames in a deterministic ~\cite{villegas2018decomposing, wichers2018hierarchical,iccv17_unsup_video, Blattmann_2021_CVPR} or stochastic manner~\cite{SAVP, franceschi2020stochastic, Castrejon_2019_ICCV, SV2P}, as pursued here. Several works 
decrease the complexity of the synthesis task by using keypoint annotations \cite{2019svrnn, keyin_kosta} 
as conditioning information.  A major drawback of this approach is the requirement of 
semantic keypoint labels which limit consideration to highly structured objects, like humans,
and thus exclude the broader range of imagery we consider, e.g., natural scenes.
Recent methods aim at improving video prediction quality by use of high capacity architectures
with high computational demands, 
operating in the latent~\cite{latent_video_transformer} or pixel-space \cite{weissenborn2020scaling}, or using attention \cite{DVDGAN}. 
In contrast, we propose a model for understanding the image-to-video synthesis process by learning a bijective transformation between the image and video domains using a dedicated residual representation.

\noindent{\bf Dynamic texture synthesis.}
Previous work has given special attention to generating dynamic textures.
This work can be divided into two groups: (i) methods that exploit the statistics of dynamics textures \cite{two-stream, learning_dynamic_Xie, stgconvnet} and (ii) learning-based approaches \cite{mdgan, dtvnet, DeepLandscape, Animating_landscapes, R3_reference_coop_training}. To generalize to other video domains, beyond dynamic textures, we introduce a learning-based approach. MDGAN \cite{mdgan} generates landscape videos from a static scene in a deterministic manner. Several methods (e.g., \cite{Animating_landscapes,dtvnet}) consider optical flow in their
video generation pipeline.
The use of optical flow limits application to specific types of 
imagery, like clouds, at the exclusion of other dynamic textures which grossly violate
standard optical flow assumptions \cite{two-stream}.
DeepLandscape \cite{DeepLandscape} extends the structure of StyleGAN~\cite{karras2019stylebased} to animate landscape images. Their model does not attempt to learn full temporal dynamics of videos and works only by a complex optimization scheme for inference, similar to \cite{gatys2015neural} for style transfer. In contrast, our approach allows for efficient feedforward image-to-video synthesis while also maintaining visual quality and temporal coherence.  

\noindent{\bf Invertible Neural Networks.} Invertible neural networks (INNs) are bjiective functions which makes them attractive for a variety of tasks, such as analyzing inverse problems \cite{ardizzone2019analyzing}, interpreting neural networks \cite{esser2020disentangling}, and representation learning \cite{jacobsen2018irevnet}. In particular, INNs can be implemented as normalizing flows \cite{rezende2016variational}, a special class of likelihood-based generative models which have recently been applied to various tasks, such as image synthesis \cite{kingma2018glow, ardizzone2019guided, pumarola2020cflow}, domain  transfer \cite{rombach2_network, rombach2020making, esser2020disentangling, wavelet_jason}, superresolution \cite{lugmayr2020srflow, wavelet_jason}, and video synthesis \cite{kumar2020videoflow}. In contrast, we use a conditional normalizing flow model to learn a dedicated residual latent, capturing information not contained in the input image.  This allows us to both more efficiently learn the bijective mapping and to consider explicit controlling factors. 

%% file: chapters/method.tex
\begin{figure*}[t]
    \centering
    \includegraphics[width=1\textwidth]{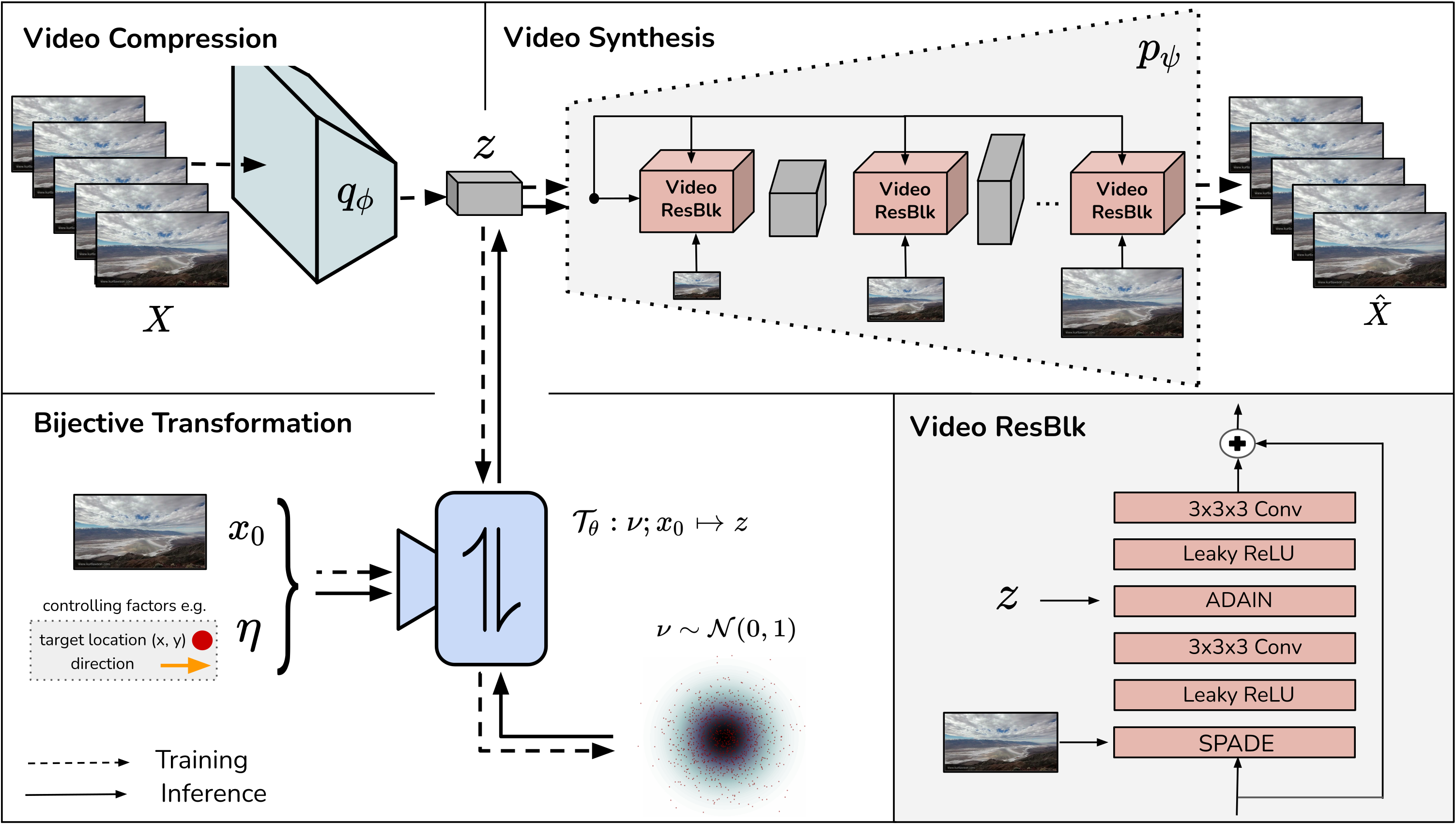}
    \caption{
    Overview of our proposed framework. We learn an information preserving video representation $z$ using our conditional generative model consisting of an encoder $q_\phi$ as well as the corresponding decoder $p_\psi$. The decoder consists of dedicated video residual blocks shown in bottom right. 
    After establishing the video representation, we learn a bijective transformation $\mathcal{T}$ conditioned on the starting frame $x_0$ and an optionally provided control factor $\eta$.
    During inference, we sample a residual $\nu$, encapsulating the scene dynamics, from the prior distribution and use $\mathcal{T}_\phi$ to obtain the video representation $z$. Using our decoder we can then synthesize novel video sequences. Training and inference is indicated by the dotted and solid lines, respectively.
    }
    \label{fig:method}
    \vspace{-0.2cm}
\end{figure*}

Our goal is to learn the interplay between images and video by explaining video in terms of a single image and the (stochastic) information not captured by the image about the video. Together the deterministic and stochastic content allow us to tackle the problem of image-to-video synthesis. In Sec.~\ref{sec:bijection}, we begin by motivating and introducing our conditional bijective framework for image-to-video mappings and Sec.~\ref{sec:residual} describes the learning process.
Sec.~\ref{sec:cVAE} presents our generative model for video synthesis operating on our learned transformation.
Finally, in Sec.~\ref{sec:control} we extend our model to directly exercise control over factors captured in the residual latent, e.g., direction of motion. Fig.~\ref{fig:method} provides an overview of our approach.

\subsection{Bijection for Image-to-Video Synthesis} 
\label{sec:bijection}
Given an initial image, $x_0 \in \mathbb{R}^{d_x}$, image-to-video synthesis generates a video sequence, $X = [ x_1,\dots,x_T ]$. This problem is inherently underdetermined with many possible videos conceivable based on $x_0$. As a result, we cannot synthesize or explain a video merely with a single frame, but require additional information, $\nu$, such as the scene dynamics. Video synthesis can then be framed as mapping $x_0$ and a residual $\nu$ onto a video $X$ or, equivalently, a representation $z$ thereof,
\begin{equation}
z = \mathcal{T}(\nu;x_0) \;.
\label{eq:imagetovideo2}
\end{equation}

Commonly, stochastic video prediction methods \cite{SAVP, franceschi2020stochastic, 2019svrnn} only focus on synthesizing \emph{arbitrary realistic} videos for a single initial or a sequence of frames. In contrast, understanding this synthesis process not only demands to explain the missing information $\nu$ to be inferred, but also to recover the residual information from video so that it can be modified subsequently. Explaining a video thus requires to estimate this residual information $\nu$, so that $x_0$ and $\nu$ together are isomorphic to the representation $z$ of the video $X$. Consequently, $\mathcal{T}$ needs to be a conditional bijective mapping between videos and their description in terms of a starting frame $x_0$ and the remaining residual information $\nu$.
\subsection{Inferring an Explicit Residual Representation}
\label{sec:residual}
Given a single frame $x_0$, a multitude of videos are possible with a corresponding $z$, 
\begin{equation}
z \sim p(z|x_0) \; .
\end{equation}
Since $\nu$ contains all the information of $z$ not captured in $x_0$ and $\mathcal{T}$ is conditionally bijective, we can invert \eqref{eq:imagetovideo2} to obtain the residual
\begin{equation}
\nu = \mathcal{T}^{-1}(z;x_0) \; .
\label{eq:backward}
\end{equation}
Then, by the change-of-variables theorem for probability distributions, $\mathcal{T}^{-1}$ transforms $p(z|x_0)$ as
\begin{align}
p( z \vert x_0) &= \frac{p(\nu \vert x_0)}{\vert \det J_{\mathcal{T}}(\nu;x_0) \vert} \\
&= p(\mathcal{T}^{-1}(z;x_0) \vert x_0) \cdot \vert \det J_{\mathcal{T}^{-1}}(z;x_0) \vert \; ,
\label{eq:transformation}
\end{align}
where $J_{\mathcal{T}}$ denotes the Jacobian of the transformation $\mathcal{T}$ and $\vert \det [\cdot] \vert $ the absolute value of the determinant of its input. 
\\
Using the transformed distribution, $p(z|x_0)$, we can now directly learn our transformation $\mathcal{T}$ and the distribution $p(\nu|x_0)$ by maximum likelihood estimation (MLE). 
To this end, we need to choose an appropriate prior distribution, which can be analytically evaluated and easily sampled. 
Since we factorize the residual information $\nu$ from the starting frame $x_0$, we can assume $p(\nu \vert x_0) = q(\nu)$ and, thus, resort to the widely used standard normal distribution $q(\nu) = \mathcal{N}(\nu \vert 0, \mathbf{1})$~\cite{VAE,yan2016cVAE,goodfellow2014GAN}. Moreover, we parametrize $\mathcal{T}$ as an invertible neural network~\cite{papamakarios2019normalizing, dinh2017density, brubaker_survey} $\mathcal{T}_\theta$ with parameters $\theta$ which, given the image $x_0$, translates between the representations $z$ and $\nu$. Thus, we arrive at the negative log-likelihood minimization problem
\begin{equation}
    \min_{\theta \in \Theta} \; \mathbb{E}_{z, x_0} \left[\log q(\mathcal{T}_\theta^{-1}(z ; x_0)) - \log \vert \det J_{\mathcal{T}_\theta^{-1}}(z ; x_0) \vert \right] \, .
\end{equation}
By simplifying using the standard normal prior and dropping resulting constant terms, we finally arrive at our final objective function
\begin{equation}
    \min_{\theta \in \Theta} \; \mathbb{E}_{z, x_0} \left[ \Vert \mathcal{T}_\theta^{-1}(z ; x_0) \Vert_2^2 -  \log \vert \det J_{\mathcal{T}_\theta^{-1}}(z ; x_0) \vert \right] \, .
\end{equation}
%
Due to the information-preserving, isomorphic mapping $\mathcal{T}_\theta$, $\nu$ indeed captures the latent information in $X$ not explained by $x_0$.

To generate a video representation $z$ based on an initial frame $x_0$, we first sample a residual representation $\nu \sim q(\nu)$ and then apply \eqref{eq:imagetovideo2} to obtain $z = \mathcal{T}_\theta(\nu; x_0)$.


\subsection{Generative Model for Video Synthesis} \label{sec:cVAE}
We now learn a decoding $p(X|z)$ to synthesize video sequences based on $z$. Since we require $z$ to be a compact, information-preserving video representation, we also need to learn the corresponding encoding $q(z|X)$. Simultaneously learning both is naturally expressed by an autoencoder~\cite{VAE}. Moreover, to optimally enable learning the transformation $\mathcal{T}_\theta$, we consider the following modelling constraints: \emph{(i)} the representation $z$ of the input should be maximally information-preserving to fully capture the residual dynamics information, \emph{(ii)} we model the residual $\nu$ to be a continuous probabilistic model, thus the bijection property of $\mathcal{T}_\theta$ requires $q(z|X)$ to be a strictly positive density, and \emph{(iii)} reducing the complexity of the representation $z$ eases the task of learning the bijective mapping $\mathcal{T}_\theta$. Thus, while still fully capturing scene dynamics in $z$, we ideally exclude all information in the video which is already present in the initial image $x_0$.

\noindent\textbf{Learning $p(X|z)$ and $q(z|X)$.} Variational latent models~\cite{VAE} are a straightforward choice for stochastic autoencoders. To address \emph{(iii)} above, we use a conditional variational autoencoder~\cite{yan2016cVAE} with a parametrized encoder $q_\phi(z|X)$ and a parametrized, conditional decoder $p_\psi(X|x_0,z)$ with $(\phi,\psi)$ being their trainable parameters. Such models encourage the distribution of information among latent variables due to the regularization of the capacity of the latent encoding~\cite{chen2016variational,zhao2017infovae, burgess2018understanding}. Thus, using $x_0$ as a conditioning to represent most of the scene content, the complexity of $z$ can be reduced by forcing the network capacity to focus on capturing the latent information in $X$. To balance this with maximally preserving the latent residual information in $X$, we introduce a weighting parameter $\beta$ to the standard variational lower bound~\cite{burgess2018understanding}, 
%
\begin{equation}
\begin{split}
\mathcal{L}_{p_\psi,q_\phi} = & \; \mathbb{E}_{z\sim q_\phi(z|X)} \left[ \log p_\psi(X| x_0, z) \right] \\
& - \beta D_{\text{KL}}(q_\phi(z|X)||q(z)) \; ,
\end{split}
\label{eq:elbo}
\end{equation}
where $q(z)$ denotes a standard normal prior on the encoder $q_\phi$. The first term optimizes the synthesis quality of the decoding process, thus maximizing information-preservation. While the second term regularizes $q_\phi(z|X)$ to match the prior $q(z)$ which constrains its capacity and, thus, encourages the distribution of information among $x_0$ and $z$ to ease subsequent learning of $\mathcal{T}_\theta$. Hence, $\beta$ allows us to directly balance the informativeness of $z$ and its complexity~\cite{chen2016variational, zhao2017infovae,burgess2018understanding}.

\noindent\textbf{Building the video synthesis model.} The design of generative architectures significantly influences their synthesis capabilities, especially when dealing with highly complex data. 
In our conditional model this particularly affects the interplay between information in $x_0$ and $z$ in $p_\phi(X|z,x_0)$. To this end, we construct the conditional decoder $p_\psi$ using a sequence of $n$ dedicated video residual blocks operating on increasing spatial and temporal feature resolutions. To optimally facilitate the interplay between $z$ and the content information in $x_0$, we combine them both in each block and, thus, at all scales of $p_\psi$.
Fig.~\ref{fig:method} illustrates the general structure of our video residual blocks used for decoding to a video. The conditioning $x_0$ is incorporated using a SPADE~\cite{spade} normalization layer to preserve semantic information throughout the generator. The video representation $z$ is added by means of an ADAIN~\cite{karras2019stylebased} layer to provide video information at all scales of the decoder. Our encoder $q_\phi$ is implemented as a 3D-ResNet~\cite{he2015ResNet} to capture the scene dynamics evolving over time in an input video.

\noindent\textbf{Overall training objective.} Following common practice~\cite{VAE}, we train our conditional model, \eqref{eq:elbo}, using an $L_1$ reconstruction loss. To emphasize perceptual quality \cite{larsen2016autoencoding} we use a frame-wise perceptual loss $\ell^{\phi}$ \cite{dosovitskiy201perceptual, johnson2016perceptual}. Similar to previous work \cite{DVDGAN, vid2vid}, we use a discriminator $\mathcal{D}_S$ applied to each frame and $\mathcal{D}_T$ on the temporal level. Both discriminators are optimized using the hinge formulation \cite{lim2017geometric, big_gan_brock}. Thus, the overall training objective can be summarized as
\begin{equation}
   \mathcal{L} = \mathcal{L}_{p_\psi,q_\phi} + \mathcal{L}_{\mathcal{D}_T} + \mathcal{L}_{\mathcal{D}_S}.
\end{equation}
%
Please see the supplemental for further details of our loss.
%

\begin{figure}[t]
    \centering
    \includegraphics[width=0.49\textwidth]{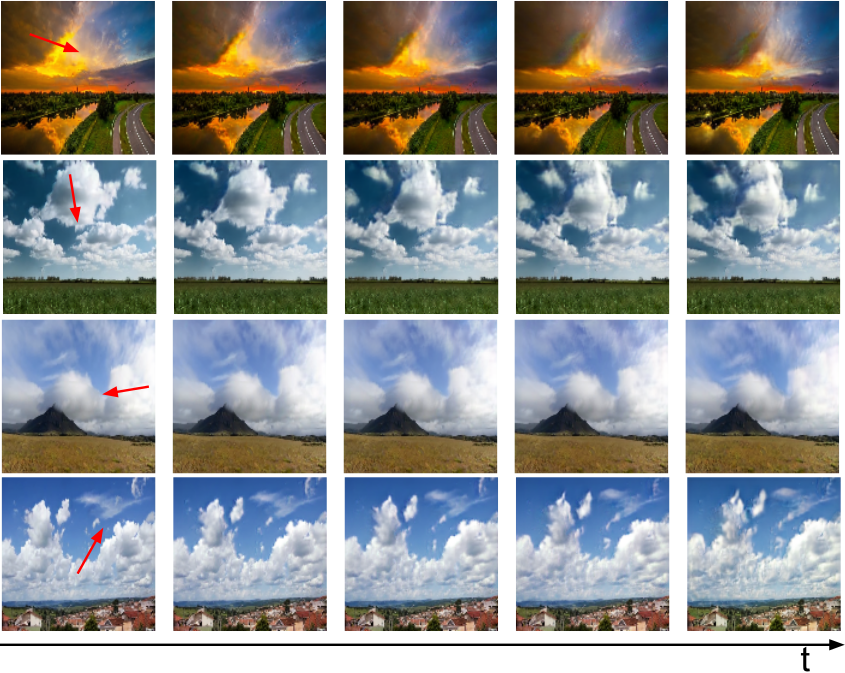}
    \caption{Stochastic video synthesis on Landscape~\cite{mdgan} showing subtle motions. Red arrows indicate the direction of motion. Best viewed as video provided in the supplemental.
    }
    \label{fig:landscape}
\end{figure}
\begin{table}[t]
\resizebox{\columnwidth}{!}{
    \begin{tabular}{l||c|c|c|c|c|c}
    \hline
    Method & LPIPS $\downarrow$ & FID $\downarrow$ & DTFVD $\downarrow$ & FVD $\downarrow$ &  DIV VGG $\uparrow$ & DIV I3D $\uparrow$\\
    \hline \hline
    MDGAN$^{2}$ \cite{mdgan} &  0.49 & 68.9 & 2.35 & 385.1 & -- & --\\
    DTVNet$^{2}$ \cite{dtvnet} &  0.35 & 74.5 & 2.78 & 693.4 & 0.00 & 0.00\\
    DL$^{2,\dagger}$~\cite{DeepLandscape} & 0.41 & 41.1 & 1.73 & 351.5 & -- & --\\
    AL$^{2}$ \cite{Animating_landscapes}  & 0.26 & 16.4 & 1.24 & 307.0 & \textbf{0.97} & 0.71 \\
    Ours  & \textbf{0.23} & \textbf{10.5} & \textbf{0.59} & \textbf{134.4} & 0.71 & \textbf{1.22} \\
    \hline
    \end{tabular}
    }
    \vspace{0.05mm}
    \caption{Quantitative evaluation 
  of video synthesis quality and diversity on Landscape \cite{mdgan}. Numeric superscripts indicate the source of the results, cf. Sec.~\ref{sec:quant:eval}. The diversity score based on the I3D~\cite{inception} trained on DTDB~\cite{dynamic_dataset} can be found in the supplemental. $^\dagger$ provided pretrained model from DL~\cite{DeepLandscape} was trained on their unreleased dataset.}
    \label{table:landscape}
    \vspace{-0.4cm}
    
\end{table}

\subsection{Controllable Video Synthesis} \label{sec:control}
%
There are many factors comprising the latent residual $\nu$. Understanding the image-to-video process allows us to directly exercise control over such factors and thus over the progression of the depicted scene in the input image $x_0$. Assuming $\eta \in \mathbb{R}^{d_\eta}$ represents such a factor, e.g., the target location of a moving object, we can explicitly model it while learning our bijective mapping $\mathcal{T}_\theta$ as $\mathcal{T}_\theta(\nu;x_0,\eta)$. Note, now $\nu$ constitutes the residual latent information to \textit{both} $x_0$ and $\eta$. Since such individual factors are typically low in information themselves, in general there is no benefit in considering them when learning the conditional decoder $p_\psi$ in contrast to the richer information in $x_0$. Image-to-video synthesis now extends to additionally manually adjusting $\eta$ to a fixed value $\eta^*$ to infer a video representation $z = \mathcal{T}_\theta(\nu;x_0,\eta^*)$ which is then used to synthesize a video sequence using $p_\psi$.
%

%% file: chapters/results.tex
\begin{figure}[t]
    \centering
    \includegraphics[width=1\linewidth]{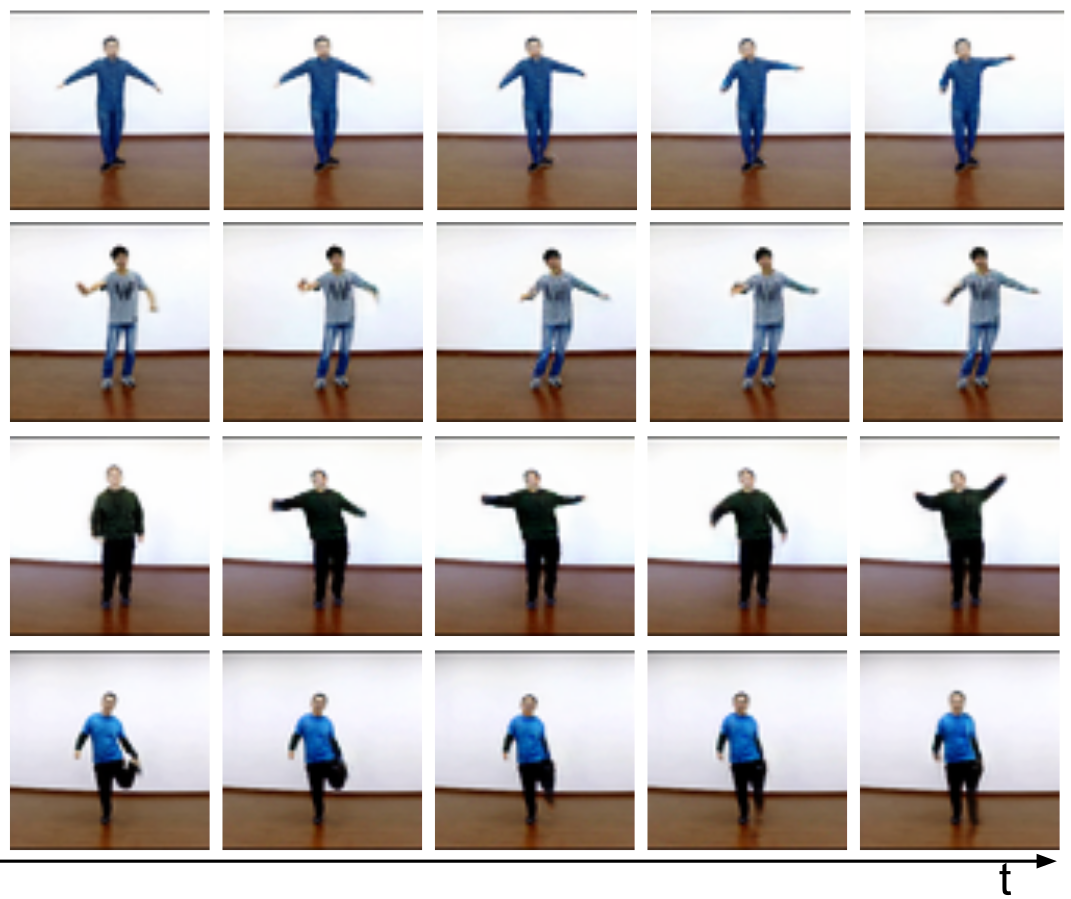}
    \caption{Stochastic video synthesis on iPER~\cite{iPER} showing structured, diverse human motion.
    Best viewed as video provided in the supplemental.
    }
    \label{fig:human}
\end{figure}

\begin{table}[t]
    \resizebox{\linewidth}{!}{
    \centering
    \begin{tabular}{l||c|c|c}
    \hline
    Method& FVD $\downarrow$& DIV VGG $\uparrow$& DIV I3D $\uparrow$ \\
    \hline \hline
    SAVP$^{3}$~\cite{SAVP} &  368.6 & 0.00$^*$ & 0.01$^*$\\
    SRVP$^{3}$~\cite{franceschi2020stochastic} & 336.3 & 0.34 & 1.01\\
    IVRNN$^{3}$~\cite{Castrejon_2019_ICCV} & 191.4 & 0.23& 0.57\\
    Ours & \textbf{132.9} & \textbf{0.50} & \textbf{1.63}\\
    \hline
    Ours w/o cINN & 180.6 & 0.32 &1.21 \\
    Ours w/o $x_0$ & 381.5 & 0.73 &2.15\\
    Ours w/o ADAIN & 156.7 &0.48 &1.60 \\
    \hline
    \end{tabular}
    }
    \vspace{0.05mm}
    \caption{Quantitative evaluation 
  of video synthesis quality and diversity on iPER~\cite{iPER}. Numeric superscripts indicate the source of the results, cf. Sec.~\ref{sec:quant:eval}. $^*$ SAVP experienced mode collapse due to training instabilities originating from the two involved discriminators. 
  }
    \label{table:human}
    \vspace{-0.4cm}
\end{table}

\begin{table*}[t]
    \resizebox{\textwidth}{!}{
    \begin{tabular}{l|| ccccc | ccccc | ccccc | ccccc}
    \hline
     \multirow{2}{*}{Method} &
     \multicolumn{5}{c}{\textbf{Fire}} & 
     \multicolumn{5}{c}{\textbf{Vegetation}} & 
      \multicolumn{5}{c}{\textbf{Waterfall}} & 
      \multicolumn{5}{c}{\textbf{Clouds}} \\
      & {LPIPS $\downarrow$} & {FID $\downarrow$} & {FVD $\downarrow$} & {DTFVD $\downarrow$} & {DIV $\uparrow$} 
      & {LPIPS $\downarrow$} & {FID $\downarrow$} & {FVD $\downarrow$} & {DTFVD $\downarrow$} & {DIV $\uparrow$} 
      & {LPIPS $\downarrow$} & {FID $\downarrow$} & {FVD $\downarrow$} & {DTFVD $\downarrow$} & {DIV $\uparrow$} 
      & {LPIPS $\downarrow$} & {FID $\downarrow$} & {FVD $\downarrow$} & {DTFVD $\downarrow$} & {DIV $\uparrow$}  \\
      \hline
      \hline
    DG$^{3}$ \cite{learning_dynamic_Xie} & 0.18 & 29.4 & 361.3 & 0.40 & -- & 0.22 & 71.6 &  290.3 & 0.86 & -- & 0.25 & 143.4 & 1680.6 & 2.41 & -- & 0.17 & 73.5 & 217.5 & 0.40 & --\\
    \hline
    AL$^{3}$ \cite{Animating_landscapes} & 0.28 & 48.4 & 1475.9 & 11.42 & 0.74 & 0.28 & 48.9 & 271.0 & 1.48 & \textbf{0.93} & 0.32 & 124.3 & 1847.8 & 5.94 & \textbf{0.98} & 0.27 & 38.7 & \textbf{142.1} & 0.76 & \textbf{1.52}\\
    Ours & \textbf{0.23} & \textbf{24.2} & \textbf{376.8} & \textbf{0.79} & \textbf{1.10} & \textbf{0.21} & \textbf{18.2} & \textbf{123.8} & \textbf{0.52} & 0.86 & \textbf{0.25} & \textbf{66.8} & \textbf{1126.5} & \textbf{2.52} & 0.61 & \textbf{0.25} & \textbf{18.3} & 179.3 & \textbf{0.73}& 0.98\\
    \hline
  \end{tabular}
 }
 \vspace{0.05mm}
  \caption{Quantitative evaluation 
  of video synthesis quality and diversity (based on VGG~\cite{simonyan2015VGG}) on DTDB \cite{dynamic_dataset}. The diversity score based on the I3D~\cite{inception} trained on DTDB~\cite{dynamic_dataset} can be found in the supplemental.
  Note, DG \cite{learning_dynamic_Xie} directly optimizes on test samples. Numeric superscripts indicate the source of the results, cf. Sec.~\ref{sec:quant:eval}.}
  \label{table:dynamic_texture}
  \vspace{-0.45cm}
\end{table*}
We evaluate the efficacy of our video synthesis method
on a diverse set of four video datasets which range from human motion to 
stochastic dynamics as encompassed by natural landscape scenery.
Video prediction results and comparisons are best viewed as videos which are available in the supplemental and on our project page\footnote{\webpage}. Implementation details can be found in the supplemental. Our PyTorch~\cite{pytorch} implementation can be found on our GitHub page\footnote{\githubpage}. Unless otherwise stated, we generate 16 frame predictions.
\subsection{Datasets}
 Here, we summarize the four diverse datasets used in our evaluation. We train all models on a sequence length of 16. A detailed description of the evaluation protocol for each dataset can be found in the supplemental.
\\
\textbf{Landscape} \cite{mdgan} consists of $\sim\hspace{-3pt}3000$ time-lapse videos of dynamic sky scenes, e.g.,  cloudy skies and night scenes with moving stars. This dataset contains a wide range of sky appearances and motion speeds. Following previous work \cite{mdgan, dtvnet}, we evaluate on a sequence length of 32 frames. We compare with recent work on landscape synthesis \cite{mdgan, Animating_landscapes, dtvnet, DeepLandscape}. To generate sequences of length 32 we apply our model sequentially, meaning we use the last predicted frame from the last generated 16 frame block as input for the next set of 16 frames. 
\\
\textbf{Dynamic Texture DataBase (DTDB)} \cite{dynamic_dataset} contains more than 10,000 dynamic texture videos. 
For evaluation, we focus on the following classes:
fire, clouds, vegetation, and waterfall. Each texture class consists of $150$ to $300$ videos. We train one model for each texture (same as for~\cite{Animating_landscapes, learning_dynamic_Xie}) on a sequence length of 16 on a resolution of $128 \times 128$.
\\
\textbf{BAIR Robot Pushing} \cite{bair} consists of a randomly moving robotic arm that pushes and grasps objects in a box. It contains around 40k training and 256 test videos. This dataset is used by prior work as a benchmark due its stochastic nature and the real-world application. We follow the standard protocol \cite{weissenborn2020scaling, FVD, DVDGAN, latent_video_transformer} and evaluate on a sequence length of 16 frames on a resolution of $64\times 64$. 
\\
\textbf{Impersonator (iPER)} \cite{iPER} is a recent dataset that contains humans with diverse styles of clothing executing various random actions. The entire dataset contains $206$ videos with a total of $241,564$ frames. We follow the train/test split defined in  \cite{iPER} which leads to training set  and test sets containing $180$k and $49$k frames, respectively. We evaluate our model on a sequence length of $16$ on a $64\times 64$ resolution.
\subsection{Evaluation Metrics}
\noindent
\textbf{Synthesis quality.} We evaluate the video synthesis quality using the Fr\'echet Video Distance (FVD) \cite{FVD} which is sensitive to both perceptual quality and temporal coherence. This metric represents the spatiotemporal counterpart to the Fr\'echet Inception Distance (FID) \cite{FID} which is based on an I3D network \cite{inception} trained on Kinetics \cite{Kinetics-original}, a large-scale human action dataset. To evaluate dynamic textures, we introduce the Dynamic Texture Fr\'echet Video Distance (DTFVD) by replacing the pre-trained network with one we trained on DTDB for classification \cite{dynamic_dataset}. The motivation behind introducing DTFVD is that we seek a metric that is sensitive to the types of dynamics encapsulated by dynamic textures, rather than human action-related motions as captured by FVD. To further evaluate dynamic textures, we also evaluate perceptual quality in terms of the FID \cite{FID} and the Learned Perceptual Image Patch Similarity (LPIPS) \cite{dosovitskiy201perceptual, johnson2016perceptual} metrics.
\\
\textbf{Diversity.}
Photorealism and plausible dynamics are not the only factors we are interested in. In addition, our model is capable of stochastically generating plausible videos from a single image. Following previous work~\cite{SAVP} on video synthesis, we measure the diversity between video sequence predictions given an initial frame $x_0$ as their average mutual distance in the feature space of a VGG-16 network \cite{simonyan2015VGG} pre-trained on ImageNet~\cite{ImageNet}. In contrast to \cite{SAVP}, we use the Euclidean distance instead of the Cosine distance. Moreover, we also report diversity on pre-trained I3D~\cite{inception} models (similar to above) which is sensitive to both appearance and motion instead of comparing samples frame-wise. We discuss and compare our chosen diversity measures in the supplemental. 
\begin{figure}[t]
    \centering
    \includegraphics[width=1\linewidth]{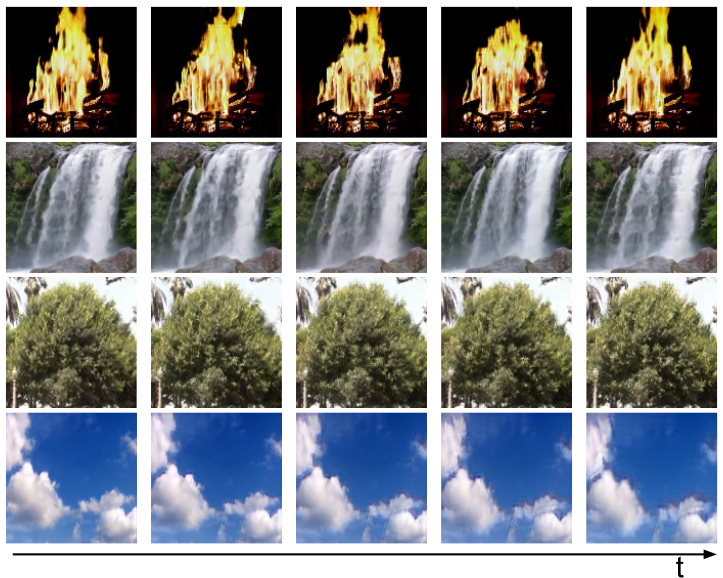}
    \caption{Stochastic video synthesis on DTDB~\cite{dynamic_dataset} for diverse texture categories. Best viewed as video provided in the supplemental.
    }
    \label{fig:DT}
    \vspace{-0.5cm}
\end{figure}
\subsection{Quantitative Evaluation}
\label{sec:quant:eval}
For comparison, we use reported performance from the corresponding paper (marked by $^{1}$), where possible, otherwise we report numbers based on pretrained models (marked by $^{2}$) or retrained models using the official code (marked by $^{3}$) provided by the author. 
\begin{table}[t]
\resizebox{\linewidth}{!}{
    \centering
    \begin{tabular}{l||c|c|c}
    \hline
    Method & FVD $\downarrow$& DIV VGG $\uparrow$&DIV I3D$\uparrow$\\
         \hline
         \hline
         Video Flow$^{1}$ \cite{kumar2020videoflow} & 131.0 & -- & --\\
         SRVP$^{2}$~\cite{franceschi2020stochastic} & 141.7 & 0.93 & 1.65\\
         IVRNN$^{3}$~\cite{Castrejon_2019_ICCV} & 121.3  & 0.69 & 1.13\\         
         SAVP$^{1,2}$~\cite{SAVP} & 116.4 & \textbf{0.98} & 1.70\\
         LVT$^{1}$  ~\cite{latent_video_transformer} & 125.8 & -- & -- \\
         DVD-GAN$^{1}$ \cite{DVDGAN} & 109.8 & -- & -- \\
         Video Transformer$^{1}$~\cite{weissenborn2020scaling} & \textbf{94.0} & -- & --\\
      Ours & 99.3 &  \textbf{0.98} & \textbf{1.93}\\
        \hline
         Ours w/o cINN & 134.5 & 0.59 & 0.94 \\
         Ours w/o $x_0$ & 272.6  & 2.40$^\dagger$ & 2.48$^\dagger$ \\
         Ours w/o ADAIN & 131.2  &  0.78 & 1.73\\
         \hline
    \end{tabular}
    }
    \vspace{0.05mm}
    \caption{Quantitative evaluation and ablation study 
  of generation quality and diversity on BAIR \cite{bair}. Numeric superscripts indicate the source of the results, cf. Sec.~\ref{sec:quant:eval}. $^\dagger$ denotes high diversity due to artifacts.}
    \label{table:bair}
    \vspace{-0.4cm}
\end{table}

\noindent\textbf{Landscape.} Tab.\ \ref{table:landscape} provides a summary of our evaluation on Landscape in terms of perceptual quality and temporal coherence. As can be seen, we generally outperform all methods across all metrics. 
Animating Landscape (AL)~\cite{Animating_landscapes} stores the motion embeddings of all training instances in their codebook and uses them to generate videos during inference. In this way, AL is able to reproduce the diversity of the training videos.
DTVNet \cite{dtvnet} does not enforce a distribution over their representation and consequently is limited to deterministic video generations. DeepLandscape~\cite{DeepLandscape} (DL) does not learn dynamics from videos, but rather uses a manually constructed set of homographies. The pretrained model provided by DL was trained on their unreleased dataset. 
In contrast, we explicitly model and learn the dynamics distribution and by that, are able to synthesize \textit{novel} dynamics to set scenes in motion.
\\
\noindent\textbf{DTDB. } We observe similar results on DTDB (Tab.~\ref{table:dynamic_texture}) on nearly all dynamic textures (fire, waterfall, and vegetation) across all perceptual quality and coherence metrics. For the clouds, AL achieves better results due to the fact that this  motion can be faithfully described by optical flow. Here, we also consider results from methods dedicated to dynamic texture synthesis \cite{learning_dynamic_Xie, stgconvnet} as strong baselines.  These
methods are not exactly comparable as they directly optimize on {\it test samples}.
We only present results for DG~\cite{learning_dynamic_Xie}, as Xie et al.\ \cite{stgconvnet} did not converge when trained on all test samples. 
\\
\noindent\textbf{BAIR. } We achieve strong results in terms of video quality (Tab.~\ref{table:bair}), even when compared with the computationally expensive transformer based approach~\cite{weissenborn2020scaling}. 
In terms of diversity, we are on par with the state-of-the-art stochastic video prediction approaches.
\\
\noindent\textbf{iPER. } The evaluation of articulated human motion is presented in Tab.~\ref{table:human}. We achieve superior results to recent approaches for video prediction \cite{franceschi2020stochastic, Castrejon_2019_ICCV, SAVP} in terms of FVD and diversity. Note, that we only condition on one frame in comparison to the baselines which use two~\cite{SAVP, Castrejon_2019_ICCV} and eight context frames~\cite{franceschi2020stochastic}. 
\begin{figure}[t]
    \centering
    \includegraphics[width=1\linewidth]{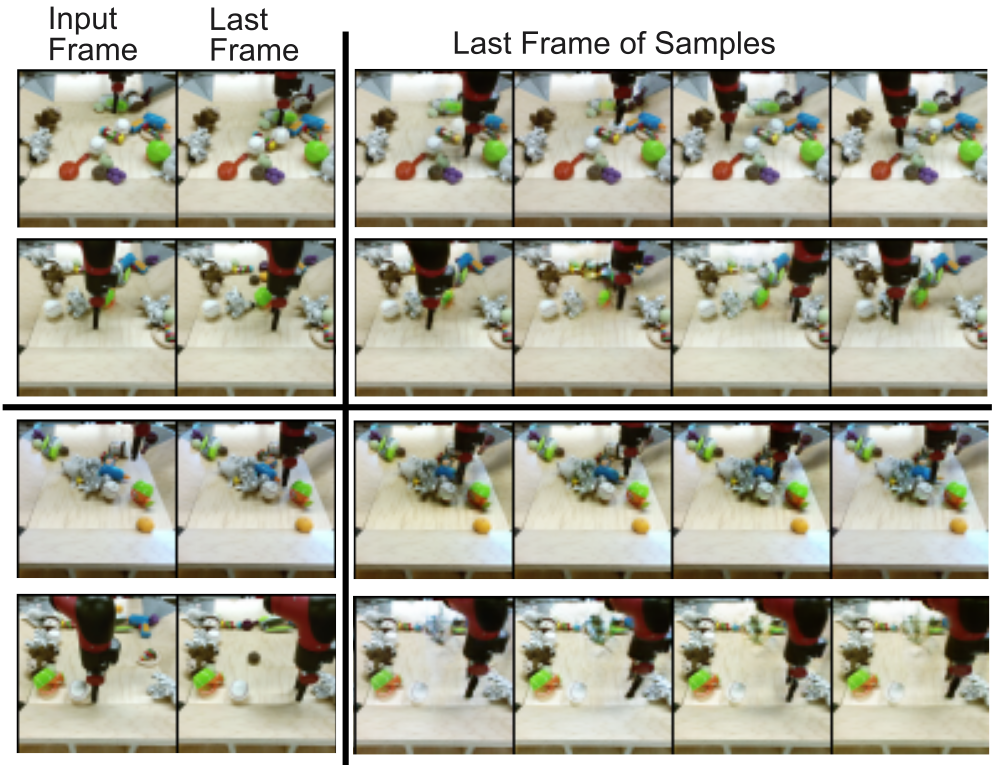}
    \caption{Qualitative evaluation of diversity on BAIR \cite{bair}.  
    {\bf Stochastic video synthesis:} (top two rows, left-to-right)  input frame  and last frame from a BAIR sequence and 
    four frames representing the last frames from sampled videos generated 
    using the input frame alone.
    The generated frames show a high degree of stochasticity in terms 
    of the end effector position, as desired.
    {\bf Controlled video synthesis:} (bottom two rows, left-to-right)  input frame and last frame from a BAIR sequence,
    and four frames representing the last frames from sampled videos
    generated using both the input frame and the 3D end effector position in the 
    last frame.
     The end effector position in the last frame is in close agreement with the position
     control input, as desired.}
    \label{fig:bair}
    \vspace{-0.4cm}
\end{figure}

\subsection{Qualitative Evaluation}
\begin{figure}[t]
    \centering
    \includegraphics[width=1\linewidth]{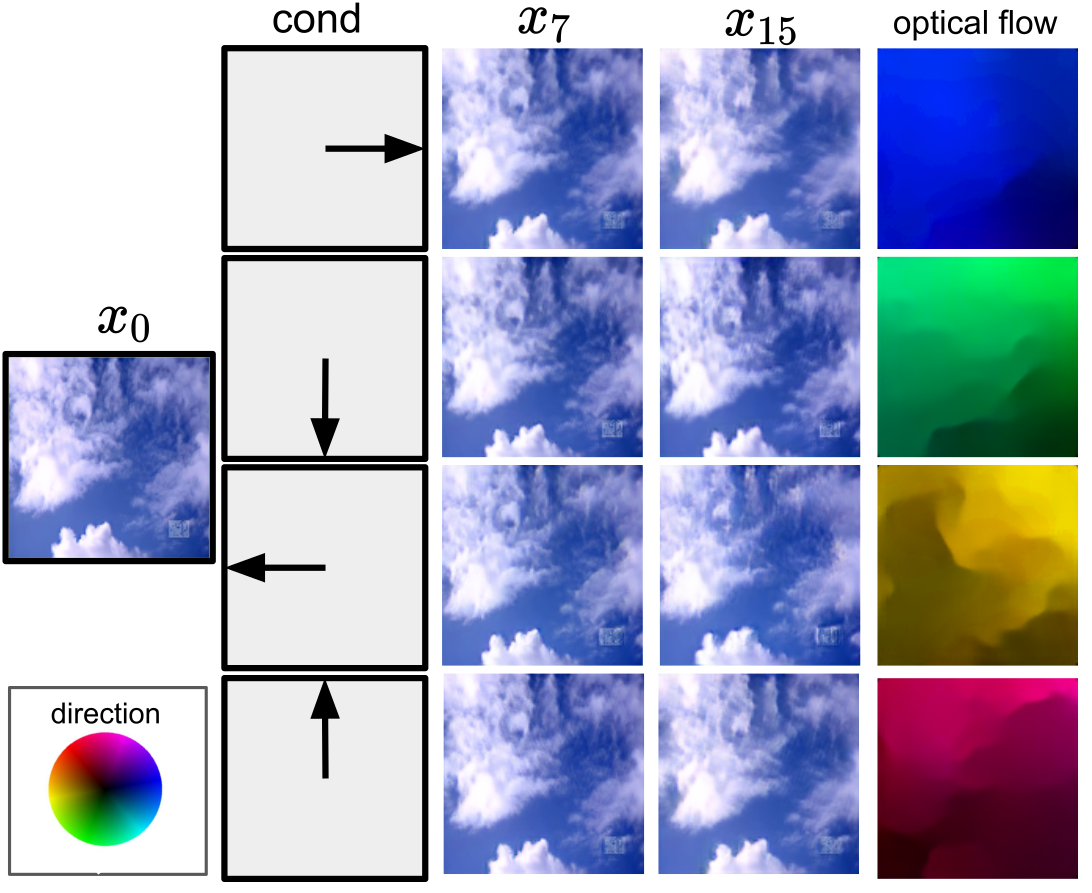}
    \caption{Examples of controlling video synthesis of clouds in DTDB \cite{dynamic_dataset} starting at frame $x_0$ using motion direction inputs (indicated by arrows). We show intermediate frames $x_7$ and  $x_{15}$. The color wheel indicates flow direction.}
    \label{fig:control_clouds}
    \vspace{-0.35cm}
\end{figure}

\begin{figure}[t]
    \centering
    \includegraphics[width=1\linewidth]{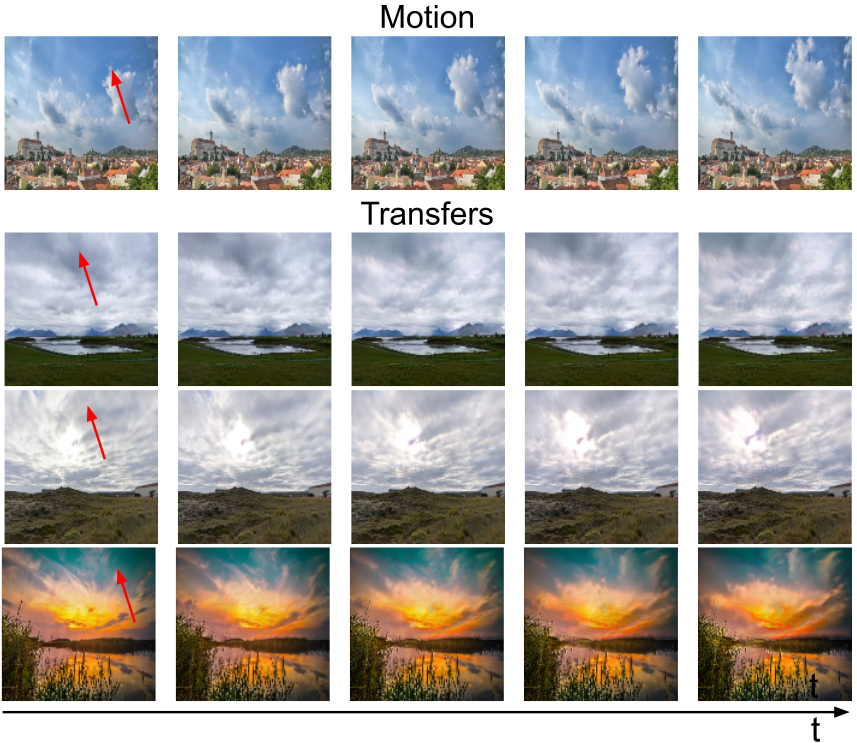}
    \caption{Transferring motion across videos on Landscape~\cite{mdgan}. (top row, left-to-right) source video for target motion. (bottom three rows, left-to-right) animating different starting frames by transferring motion from source video. 
    Red arrows indicate the 2D direction of motion.
    Best viewed as video provided in the supplemental. 
    }
    \label{fig:landscape_transfer}
    \vspace{-0.35cm}
\end{figure}

\noindent\textbf{Image-to-video synthesis.}
We provide samples for all datasets. On Landscape \cite{mdgan} we see that our model is able to synthesize realistic samples (see Fig.~\ref{fig:landscape}) from diverse, complex scenes captured in the input image.  
In Fig.\ \ref{fig:human}, we show samples on iPER~\cite{iPER} which illustrates the complexity of motion in the dataset. In Fig.\ \ref{fig:DT}, we visualize one sample per DTDB class which shows the variety of dynamic textures used for evaluation. Lastly, Fig.\ \ref{fig:bair} (top two rows) show the diversity in our video samples by way of the differences across the last generated frame per sample on BAIR~\cite{bair}.
\\
\textbf{Controllable video synthesis.}
A strength of our model is the ability to exert explicit control over the synthesis process. 
As described in Sec.\ \ref{sec:control}, we control this process by introducing a factor $\eta$.  Here, we consider two different factors for controllable video synthesis on BAIR \cite{bair} and DTDB \cite{dynamic_dataset}.
On BAIR we condition the synthesis process on the 3D location of the robot arm's end effector in the last frame;
we use the location provided in the groundtruth.  Fig.\  \ref{fig:bair} (bottom two rows)
shows several  samples of the last frame of each sequence of our controllable synthesis.
It can clearly be seen that the last frames of our samples match closely to the groundtruth end frame.
As a second example, this time on DTDB \cite{dynamic_dataset}, we condition the video synthesis of clouds based on the 2D direction of motion, again
through manipulating $\eta$.
This is visualized in Fig. \ref{fig:control_clouds} where four different directions are considered.
To aid in the visualization, we also include the optical flow fields, 
estimated with \cite{ilg2016flownet}, to show the consistency between the motion direction
used for conditioning and the direction realized in the generated videos.
As can be seen, the conditioning and generated motion directions are in close agreement. 
For results on controlled video-to-video synthesis (cf.\ Sec.~\ref{sec:control}), please refer to the supplemental.
\\
\textbf{Motion transfer}. 
Finally, we illustrate the capability of our model to transfer a motion contained in
one sequence to a set of initial frames for video synthesis. Fig.\ \ref{fig:landscape_transfer}
illustrates this process using Landscape \cite{mdgan}, where the top row contains the motion to be transferred
and the bottom three rows show the generated video sequences realized by combining
the transferred motion and the initial frames.
As can be clearly seen, the original motion is successfully transferred to each of the
scenes. 
%
\\

\vspace{-0.3cm}
\subsection{Ablation study}
To evaluate the design choices of our approach, we now perform ablation studies on BAIR~\cite{bair} and iPER~\cite{iPER}: (Ours w/o $x_0$) represents implementing our video generator, $p_\psi$, without conditioning on the input image, $x_0$, thus $z$ also captures the full scene content information, (Ours w/o ADAIN) similarly denotes removing the ADAIN input of $z$ in our proposed Video ResBlk, i.e., $p_\psi$ only has access to $z$ via the bottleneck and (Ours w/o cINN) stands for removing the cINN resulting in a cVAE framework.

In Tab.~\ref{table:human} and Tab.~\ref{table:bair}, we observe significant performance drops for all ablations compared to our full model (Ours). In particular removing the conditioning image, $x_0$, from the generator, $p_\psi$, greatly affects the synthesis quality. 
This is due to the generator not having direct access to the static information depicted in the initial frame $x_0$.
When removing the ADAIN input of $z$ from our Video ResBlk, the information of $z$ is now only available at the lowest scale of $p_\psi$, in contrast to the multi-scale information flow in our full model. Moreover, the cVAE-only model (w/o cINN) results in worse performance both in quality and diversity, which can be explained by the trade-off between synthesis quality and capacity regularization, as discussed in Sec.~\ref{sec:cVAE}.

%% file: chapters/conclusion.tex
In summary, we introduced a novel model for understanding image-to-video synthesis based on a bijective transformation, instantiated as a cINN, between the video and image domains plus residual information.
The probabilistic residual representation allows to sample and synthesize novel, plausible progressions in video with the same initial frame.
Moreover, our framework allows for incorporating additional controlling factors to guide the image-to-video synthesis process. 
Our empirical evaluation and comparison to strong baselines on four diverse video datasets demonstrated the efficacy of our stochastic image-to-video synthesis approach.

\subsection*{Acknowledgement}
This work was started as part of M.D.’s internship at Ryerson University and was supported in part by the DAAD scholarship, the NSERC Discovery Grant program (K.G.D.), the German Research Foundation (DFG) within project 421703927 (B.O.) and the BW Stiftung (B.O.). K.G.D. contributed to this work in his capacity as an Associate Professor at Ryerson University.

%% file: supplementary.tex
\section{Additional Visualizations}
For each of our experiments conducted in the main paper, we provide additional video material, consisting of 17 videos in total. To further highlight the benefits of our proposed framework, in the course of our supplemental video material, we compare to \textit{five} approaches. \textbf{Due to the collective large size of the videos, the supplemental with the corresponding videos is provided on our \href{https://bit.ly/3dg90fV}{project page}}. For each video, multiple cycles are shown (indicated left-bottom) as well as the corresponding video playback rate in frames-per-second (FPS) (right-bottom). The file structure of our provided video material is as follows:
\begin{verbatim}
  supplemental_material_222 
      |
      +--A1-Landscape
      |
      +--A2-iPER
      |
      +--A3-DTDB
      |
      +--A4-BAIR
      |
      +--A5-Controllable_Video_Synthesis
      |
      +--A6-Failure_Cases
\end{verbatim}
We next discuss the video material for each experiment individually. Each subsection matches its corresponding file (e.g., `A.1.Landscape' corresponds to `\verb+...--A1-Landscape+') which contains the discussed video sequences.

\subsection{Landscape}
\label{sec:appendix_landscape}
For the Landscape dataset \cite{mdgan}, we provide the corresponding video (\verb+Landscape_samples.mp4+) to the samples depicted in Fig.\ 3 in the main paper. Additionally, we show a qualitative comparison to previous work, i.e., AL~\cite{Animating_landscapes}, DTVNet~\cite{dtvnet}, and MDGAN~\cite{mdgan} in \verb+Landscape_comparison.mp4+, with `GT' denoting the ground-truth. We clearly observe that our model synthesizes more appealing and realistic video sequences compared to the the competing methods. Both MDGAN~\cite{mdgan} and DTVNet~\cite{dtvnet} produce blurry videos when using the officially provided pretrained weights and code from the respective webpages. While AL produces decent animations in the presence of small motion, when animating fast motions, however, warping artifacts are present, cf.\ e.g., row 3. These artifacts become even more evident when AL is applied to DTDB (Sec.~\ref{sec:appendix_dtdb}). In contrast, our method produces realistic looking results in the case of both small and large motions.
Next, we evaluate the diversity of the generated samples in \verb+Landscape_diversity.mp4+. The video contains multiple future progressions for a given starting frame, $x_0$. It can be seen that our approach produces diverse samples capturing a broad range of motion directions, as well as speeds. 
Moreover, we demonstrate in \verb+Landscape_longer_duration.mp4+ the capability of our model to synthesize longer sequences (48 frames) by sequentially applying our model on the last frame of the previously predicted video sequence.
%
%
%
%
\subsection{iPER} 
For the iPER dataset~\cite{iPER}, we provide the corresponding video (\verb+iPER_samples.mp4+) to the samples depicted in Fig.~4 in the main paper. We further provide a qualitative comparison to the best performing method IVRNN~\cite{Castrejon_2019_ICCV} on iPER in \verb+iPER_comparison.mp4+ with `GT' denoting the ground-truth. Our method produces more natural motions, e.g., row 3, compared to~\cite{Castrejon_2019_ICCV}. Note, that both methods suffer from artifacts due to the low image resolution of $64\times 64$, such as vanishing hands in motion.
%
%
\subsection{DTDB} 
\label{sec:appendix_dtdb}
For each dynamic texture from DTDB~\cite{dynamic_dataset} used in our main paper, we provide examples (\verb+Clouds.mp4+, \verb+Fire.mp4+, \verb+vegetation.mp4+, \verb+Waterfall.mp4+) for stochastic image-to-video synthesis for random starting frames, $x_0$, comparing our proposed approach to AL~\cite{Animating_landscapes} and DG~\cite{learning_dynamic_Xie}. As described in the main paper, DG~\cite{learning_dynamic_Xie} is directly optimized on test samples, thus overfitting directly to the test distribution. Consequently, we observe that their generations almost perfectly reproduce the ground-truth motion which is most evident for the clouds texture. However, their method suffers from blurring due to optimization using an L2 pixel loss. Similar to the comparisons on the Landscape dataset (Sec.~\ref{sec:appendix_landscape}), AL~\cite{Animating_landscapes} has problems with learning and reproducing the motion of dynamic textures exhibiting rapid motion changes, such as fire. This is explained by the susceptibility of optical flow to inaccuracies when capturing very fast motion, as well as dynamic patterns outside the scope of optical flow, e.g., flicker. Moreover, in the clouds examples (last row) AL wrongly sets the landscape into motion. Our model, on the other hand, produces sharp video sequences with realistic looking motions for \textit{all} textures. 

\subsection{BAIR} 
In \verb+BAIR_comparison.mp4+, we provide a qualitative comparison to a strong baseline, IVRNN \cite{Castrejon_2019_ICCV}, on the BAIR dataset~\cite{bair}. While both approaches are able to render the robot's end effector and the visible environment well, we observe significant differences when it comes to the effector interacting with or occluding background objects. An example of this difficulty can be seen when interacting with the object in the middle of the scene in row 2. IVRNN is unable to depict the object structure and texture during the interaction which results in heavy blur due to averaging over all possible future states. In contrast, this interaction looks much more natural in the video sequence predicted by our model (also row 2). Moreover, the last row (back of the scene, right) illustrates a problem of IVRNN which sometimes occurs in the presence of object occlusions. Specifically, the object which is occluded at the beginning is eventually revealed and is synthesized as a blurry texture, by that, averaging over all possible realizations. Again, our model does not suffer from this problem and correctly handles object occlusions. Additionally, \verb+BAIR_diversity.mp4+ qualitatively illustrates the prediction diversity of our model by animating a fixed starting frame $x_0$ multiple times. Again, `GT' denotes ground-truth. Our model synthesizes diverse samples by broadly covering motions in the $x$, $y$, and $z$ directions.

\subsection{Controllable Video Synthesis}
In this section, we present qualitative experiments for the following controlled video prediction task: \textit{controlled image-to-video synthesis}, \textit{motion transfer}, and \textit{controlled video-to-video synthesis}.
\\
\noindent \textbf{Controlled image-to-video synthesis.} The video \verb+Endpoint_BAIR.mp4+ illustrates several image-to-video generations while controlling $\eta = (x,y,z)$, the 3D end effector position, similar to Fig.~6 in our main paper. It shows that, while in each example the effector approximately stops at the provided end position (end frame of GT), its movements between the starting and end frame, which are inferred by the sampled residual representations $\nu \sim q(\nu)$, exhibit significantly varying and natural progressions. Moreover, in \verb+Direction_Clouds1.mp4+ we provide additional video examples for controlling the direction of cloud movements with $\eta$, similar to Fig. 7 in our main paper. We observe that our model renders crisp future progressions (row 2-5) of a given starting frame $x_0$, while following our provided movement control (top row).
\\
\textbf{Motion transfer.} Next, we analyze the application of our model for the task of directly transferring a query motion extracted from a given landscape video $\tilde{X}$ to a random starting frame $x_0$. To this end, we extract the residual representation $\tilde{\nu}$ of $\tilde{X}_0$ by first obtaining its video representation $\tilde{z} = q(z|\tilde{X})$ and corresponding residual $\tilde{\nu} = \mathcal{T}_\theta^{-1}(\tilde{z};\tilde{x}_0)$ with $\tilde{x}_0$ being the starting frame of $\tilde{X}$. We use $\tilde{\nu}$ to animate the starting frame $x_0$. \verb+Transfer_Landscape.mp4+ shows that our model accurately transfers the query motion, e.g., as the corresponding direction and speed of the clouds, to the target landscape images (rows 1-3, left-to-right).
\\
\textbf{Controlled video-to-video synthesis.} 
In controlled video-to-video synthesis, we explicitly adjust the initial factor $\tilde{\eta}$ of an observed video sequence $\tilde{X}$. To this end, we first obtain its video representation $\tilde{z} = q_\phi(z|\tilde{X})$ followed by extracting the corresponding residual information $\tilde{\nu} = \mathcal{T}_\theta^{-1}(\tilde{z};\tilde{x}_0, \tilde{\eta})$. Subsequently, to generate the video sequence depicting our controlled adjustment of $\tilde{X}$, we simply choose a new value $\tilde{\eta}=\tilde{\eta}^*$ and perform the image-to-sequence inference process. This can be seen in the video \verb+Direction_Clouds2.mp4+ using cloud video sequences from DTDB \cite{dynamic_dataset}. In each example (second row), the motion direction of the query video (leftmost) is adjusted by the provided control (top row). To highlight that the residual representations $\nu$ in these cases actually correspond to the query video, we additionally animate the initial image of the query videos by sampling a new residual representation $\nu \sim q(\nu)$ and apply the same controls (bottom rows). We observe that, while the directions of the synthesized videos are identical, their speeds are significantly different, as desired. In the case of video-to-video synthesis, the movement speed remains the same, in contrast to the image-to-video case, where the movement speed has changed due to the changed residual representation.
\subsection{Failure Cases}
We highlight two types of failure cases we observed which are visualized in the video \verb+Failure_cases.mp4+:

\begin{itemize}
    \item When the starting frame depicts a complex posture (e.g., folded arms or a leg in the air) on iPER~\cite{iPER} the model has difficulty synthesizing realistic continuations.
    \item While the Landscape dataset~\cite{mdgan} mainly covers naturally progressing cloud motions, there is also a small subset of fast timelapse videos. Due to the underrepresentation of such examples in the dataset, our model struggles to correctly capture fast paced timelapse data without explicitly resorting to data-balancing techniques during training.
\end{itemize}

\section{Implementation Details}
Here, we provide a detailed overview of our network architecture as well as the training procedure. The PyTorch~\cite{pytorch} implementation can be found on our GitHub page\footnote{\githubpage}.
\subsection{Network Details}
\noindent \textbf{Encoder.} The encoder $q_\phi(z|X)$ follows the structure of a 3D ResNet-18 \cite{he2015ResNet} using GroupNorm \cite{wu2018group} as a normalization layer. Two convolutions with a kernel size of $4\times4$ are used to obtain an one-dimensional latent representation for representing the mean $\mu$ and log variance $\log \sigma ^2$. During training, we sample from $q_\phi(z|X)$ using the the reparametrization trick \cite{VAE, rezende2014stochastic}.

\begin{figure}
    \centering
    \includegraphics[width=1\linewidth]{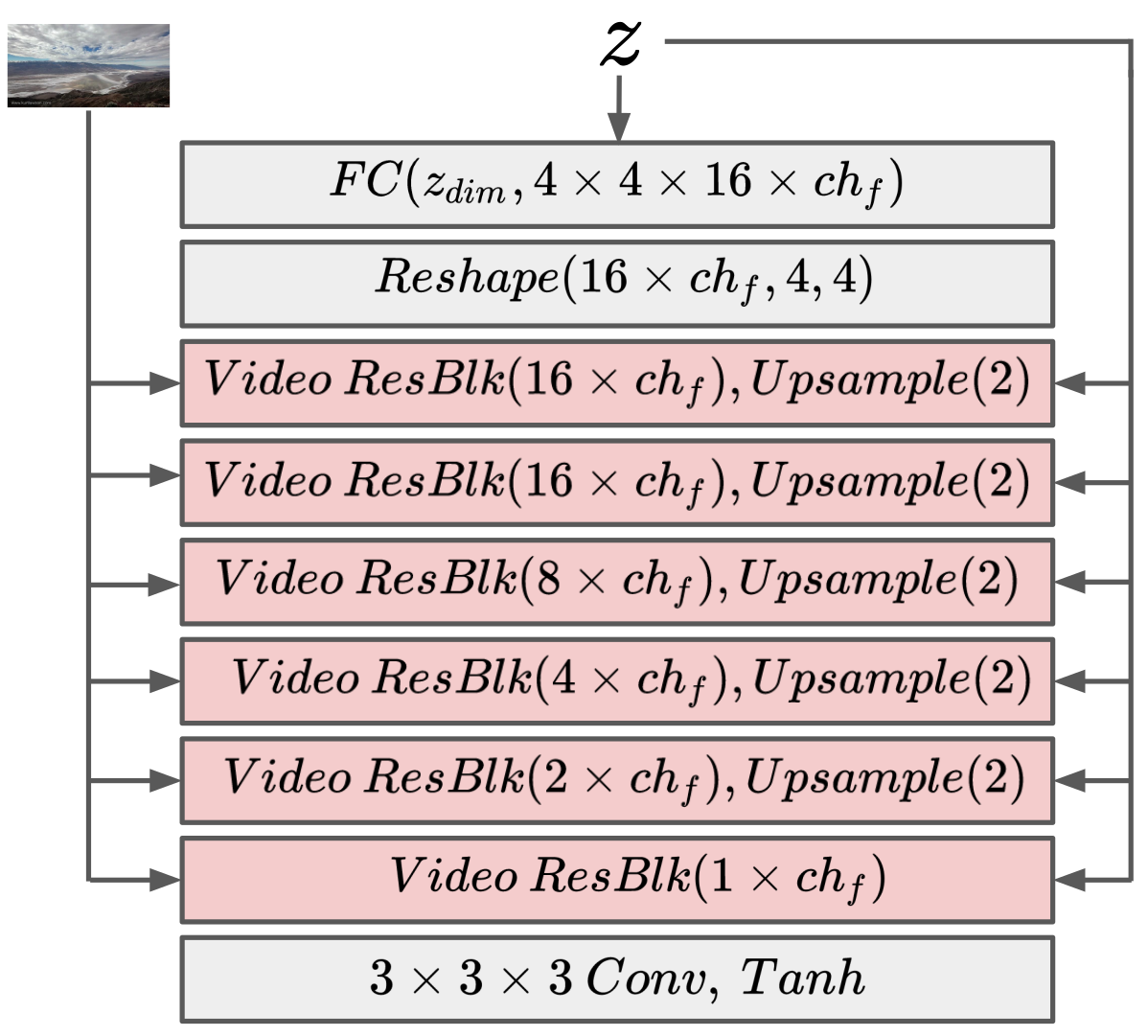}
    \caption{Overview of the decoder structure.}
    \label{fig:architecture_decoder}
\end{figure}

\noindent \textbf{Decoder.}
The decoder $p_\psi(X|x_0,z)$ consists of $n=6$ video residual blocks, with each block followed by nearest-neighbor upsampling to upscale the feature map in space and time (except the last one). This structure is illustrated in Fig.~\ref{fig:architecture_decoder}. The video representation, $z$, is inserted into the generator using a fully connected layer matching the initial feature map. The channel factor, $ch_f$, defines the number of channels and by that, the depth of the model. For BAIR and iPER, we set $ch_f$ to $64$, otherwise we set it to $32$. Depending on the dataset, time length, and resolution, the last two up-scaling layers needs to be adjusted. The video representation $z$ is inserted to the decoder using a fully connected layer matching the initial feature map. We use GroupNorm \cite{wu2018group} in SPADE~\cite{spade} and instance normalization in the ADAIN~\cite{karras2019stylebased} layer. If the input and the output channels do not match, a $1\times 1$ convolution is used to adjust the channel dimensions. For matching the output channels, we use a 3D convolution followed by a Tanh activation function. Moreover, spectral norm \cite{miyato2018spectral} is used in the decoder.

\noindent \textbf{Bijective Transformation.}
The bijective transformation, $\mathcal{T}_\theta$, is realized as a normalizing flow consisting of a stacked sequence of $n_f$ invertible neural networks (INNs) operating on the video representation, $z$. We use $n_f=20$ invertible blocks for all datasets. Each block consists $8$ normalizing flows consisting of actnorm~\cite{kingma2018glow}, affine coupling layers~\cite{dinh2017density}, and fixed shuffling layers, following previous work \cite{rombach2020making}. Each affine coupling layer is parameterized by two fully connected layers. In every affine coupling layer, we additionally insert the conditioning information following previous work~\cite{ardizzone2019guided, rombach2020making}. The feature representation for the starting frame $x_0$ is obtained by a pretrained Autoencoder optimized for reconstructing images.

\noindent \textbf{Discriminators.}
For the static discriminator, a patch discriminator \cite{pix2pix} is used and for the temporal discriminator a 3D ResNet \cite{he2015ResNet}.

\subsection{Training Details}
The loss objective for the generative model of a video sequence $X = [ x_1,\dots,x_T ] \sim p_X(X) \in \mathbb{R}^{d_X}$ with the corresponding starting frame $x_0 \in \mathbb{R}^{d_x}$ and a video representation $z \sim q_\phi (z|X) \in \mathbb{R}^{d_z}$ can be written as
\begin{equation}
\begin{split}
    \mathcal{L}_{p_\psi,q_\phi} = & \mathbb{E}_{\tiny{\substack{X \sim p_X(X) \\ z \sim q_\phi (z|X)}}} \bigg [ \lambda[\parallel X - p_\psi(X|x_0, z) \parallel_1 \\
    &+ \ell^{\phi}(X, p_\psi(X|x_0, z)) ]  - \mathcal{D}_T(p_\psi(X|x_0, z)) \\ 
    &- \mathcal{D}_S(p_\psi(X|x_0, z)) + \lambda_{F} \ell_{F}(X, p_\psi(X|x_0, z)) \bigg] \\
    &+ \beta D_{\text{KL}}(q_\phi(z|X)||q(z)) \; ,
\end{split}
\label{eq:loss_decoder}
\end{equation}
where $\ell_{F}$ denotes the feature matching loss \cite{high_res_image_syn} to stabilize the training. The hyperparameters $\lambda$ and $\lambda_F$ are both set to $10$.
\\
\indent
The loss objective for the temporal discriminator can be written as 
\begin{equation}
\begin{split}
    \mathcal{L}_{\mathcal{D}_T} &= \mathbb{E}_{\tiny{X \sim p_X(X)}} \left [ \rho (1-\mathcal{D}_T(X)) + \lambda_{GP} \parallel \nabla \mathcal{D}_T(X) \parallel_2^2 \right ] \\
    &+ \displaystyle \mathbb{E}_{\tiny{\substack{X \sim p_X(X) \\ z \sim q_\phi (z|X)}}} [\rho (1 + \mathcal{D}_T(p_\psi(X|x_0, z))],
\end{split}
\end{equation}
where $\parallel \nabla \mathcal{D}_T(X) \parallel_2^2$ denotes the gradient penalty \cite{MeschederICML2018, gulrajani2017improved} to stabilize the discriminator training and $\rho$ the ReLU activation function. The weighting factor $\lambda_{GP}$ was set to 10.
\\
\indent
For the spatial discriminator, the objective can be formulated as 
\begin{equation}
\begin{split}
    \mathcal{L}_{\mathcal{D}_S} &= \mathbb{E}_{\tiny{X \sim p_X(X)}} [\rho (1-\mathcal{D}_S(X)] \\
    &+ \displaystyle \mathbb{E}_{\tiny{\substack{X \sim p_X(X) \\ z \sim q_\phi (z|X)}}} [\rho (1 + \mathcal{D}_S(p_\psi(X|x_0, z))] .
\end{split}
\end{equation}
%
The overall loss objective can be summarized as 
\begin{equation}
   \mathcal{L} = \mathcal{L}_{p_\psi,q_\phi} + \mathcal{L}_{\mathcal{D}_T} + \mathcal{L}_{\mathcal{D}_S}.
\end{equation}
Our video synthesis model is trained using Adam \cite{kingma2017adam} with a learning rate of $2 \cdot 10^{-4}$, $\beta_1=0.5$, $\beta_2 = 0.9$, weight decay of $10^{-5}$, and exponential learning rate decay.
%
%
%
The dimension of $z$ is set to $d_z=64$ for all datasets. The weighting term $\beta$ of the Kullback-Leibler divergence loss $D_{\text{KL}}$ is set to $\beta = 1 \cdot 10^{-5}$. For the controllable video synthesis task, we discretize the conditioning $\nu_1$ to one-hot vectors. For the 3D end effector position, the x, y and z axis is discretized into 10 bins. For the clouds, the motion direction is discretized into 36 bins.
The 3D end effector position was provided by \cite{bair} and for the clouds \cite{dynamic_dataset} we manually labelled the direction. The normalizing flow, $\mathcal{T}_\theta$, was trained using Adam~\cite{kingma2017adam} with a learning rate of $1 \cdot 10^{-5}$ and linear learning rate decay.
\section{Evaluation Details}
\subsection{Diversity Metric}
Besides synthesis quality, diversity is the main criteria we use to evaluate and compare stochastic video synthesis approaches. The assessment of diversity is typically based on measures utilizing feature representations of pretrained models~\cite{SAVP, zhu2018multimodal}. For instance, SAVP~\cite{SAVP} uses a VGG network~\cite{simonyan2015VGG} trained for classification on ImageNet~\cite{ImageNet} to yield frame-wise representations of video sequences. Based on these representations, videos are compared based on their frame-wise differences measured using a given distance metric. The guiding intuition is that more diverse sample sets should exhibit larger feature differences on average. To this end, SAVP~\cite{SAVP} uses the Cosine distance. We argue that this evaluation distance has a major drawback: the Cosine distance only measures the angle between feature vectors, thus discarding crucial information represented by the vector norms. For instance, two data points may lie approximately on a line (i.e., a Cosine distance of $0$) but still are located far from each other. Hence, diversity is measured based on incomplete information. 
\begin{table}[t]
\resizebox{\linewidth}{!}{
    \centering
    \begin{tabular}{l||c|c|c|c|c}
    \hline
    Method & Landscape & Fire & Vegetation & Waterfall & Clouds\\
    \hline \hline
    AL\cite{Animating_landscapes}& 1.49 & 0.36 & 0.30 & 0.80 & 1.22 \\
    Ours & \textbf{3.41} & \textbf{1.42} & \textbf{0.98} & \textbf{1.11} & \textbf{1.51}  \\
    \hline
    \end{tabular}
    }
    \caption{Diversity scores based on the I3D~\cite{inception} trained on DTDB~\cite{dynamic_dataset}. The average difference between ground-truth samples are a factor of $\sim$ 1000 smaller for the I3D~\cite{inception} network trained on DTDB~\cite{dynamic_dataset} as the one trained Kinetics~\cite{Kinetics-original}. For presentation purposes, the numbers in the table have been multiplied by a factor of 1000. 
    }
    \label{table:div_missing_main_paper}
\end{table}
\\
\indent
To circumvent this issue, we replace the Cosine distance with the Euclidean distance which also takes the magnitude of a vector into account. Moreover, to explicitly capture temporal information, we also investigate replacing the frame-based VGG feature extractor with an I3D model~\cite{inception} which directly yields representations that capture the appearance and dynamics of the entire video sequence. Tab.~\ref{table:div_metric_comparison} compares the discussed diversity measures. It can be seen that independent of the diversity measure, the order of the approaches is the same.
We employ both VGG MSE and I3D MSE measures in our experiments. Note that the I3D feature extractors have been trained on similar datasets as the videos to be evaluated, i.e., Kinetics~\cite{Kinetics-original} for human motion~\cite{iPER} and DTDB~\cite{dynamic_dataset} for Landscape~\cite{mdgan}.  
\begin{table}[t]
\resizebox{\linewidth}{!}{
    \centering
    \begin{tabular}{l||c|c|c}
    \hline
    Method & VGG Cosine & VGG MSE & I3D MSE\\
    \hline \hline
    SAVP$^{\dagger, 3}$~\cite{SAVP} & 0.000 & 0.00 & 0.01 \\
    SRVP$^{3}$~\cite{franceschi2020stochastic} & 0.040 & 0.34 & 1.01 \\
    IVRNN$^{3}$~\cite{Castrejon_2019_ICCV}  & 0.023 & 0.23 & 0.57\\
    Ours & 0.076 & 0.50 & 1.63\\
    \hline
    \end{tabular}
    }
    \caption{Comparison of different diversity metrics on iPER~\cite{iPER}. 
    $^\dagger$ SAVP experienced mode collapse due to training instabilities originating from the two involved discriminators. The VGG based feature extractors have been pretrained on ImageNet~\cite{ImageNet}. The I3D feature extractor has been pretrained on Kinetics~\cite{kinetics}. $^3$ denotes models trained using the official code from their corresponding webpages.}
    \label{table:div_metric_comparison}
\end{table}
\indent Moreover, we report the missing diversity scores based on the I3D~\cite{inception} from the main paper on Landscape~\cite{Animating_landscapes} and DTDB~\cite{dynamic_dataset} in Tab.~\ref{table:div_missing_main_paper}.

\subsection{Evaluation Protocol}
For comparisons on each dataset, we use the reported numbers from the corresponding paper, where possible, otherwise we use pretrained models or train models from scratch using the code from the official 
webpage\footnote{
\begin{itemize}
\item[]https://github.com/edouardelasalles/srvp
\item[]https://github.com/facebookresearch/improved\_vrnn
\item[]https://github.com/alexlee-gk/video\_prediction
\item[]https://github.com/jianwen-xie/Dynamic\_ generator
\item[]https://github.com/zilongzheng/STGConvNet
\item[]https://github.com/endo-yuki-t/Animating-Landscape
\item[]https://github.com/zhangzjn/DTVNet
\item[]https://github.com/weixiong-ur/mdgan
\end{itemize}
}.
Here, we list the evaluation protocol for each dataset.
\\
\textbf{BAIR \cite{bair}.} We follow the standard protocol \cite{FVD} for computing the FVD score by evaluating videos on a sequence length of 16 on a resolution of $64 \times 64$ using all 256 test videos. Diversity is measured by predicting five future progression given the starting frames from all $256$ test sequences and computing the Euclidean distance in the VGG-16 \cite{simonyan2015VGG} as well as in the I3D~\cite{inception} feature space between the corresponding generated videos.
\\
\textbf{iPER \cite{iPER}.} For evaluating the FVD score, we use 1000 randomly sampled sequences from the test set as well as the corresponding generations. Note, for a fair comparison, we concatenate the \textit{last} conditioning frame to the generated rather than all conditioning frames since previous work condition on up to eight frames. This results in a sequence of length 17 for computing the FVD score. For computing the diversity, we predict five future progression for each of the 1000 test sequences and measure the diversity based on that.
\\
\textbf{Landscape \cite{mdgan}.} We create an evaluation set by randomly sampling six times sequences of length 32 from each test video with length over 32 resulting in 918 videos. Based on these sequences, FVD, DTFVD, LPIPS, and FID are computed. As explained in the main paper, our model is trained on a sequence length of 16 but applied two times by using the last predicted frame as input for the next prediction. For diversity, we again generate five future progressions for each sequence of the 918 evaluation sequences and use the same procedure described for BAIR.
\\
\textbf{DTDB \cite{dynamic_dataset}}. We create an evaluation set by using five sequences of length 16 from each test video resulting in between $90$ and $385$ test sequences depending on the texture.
Based on these sequences, the FVD, DTFVD, LPIPS, and FID are computed. This evaluation procedure is the same for each texture. We train one model for AL \cite{Animating_landscapes} as well as for our approach on each texture. For diversity, we again generate five future progressions for each sequence of the evaluation set and use the same procedure described for BAIR.
\subsection{Dynamic Texture FVD (DTFVD)}
\begin{figure*}
    \centering
    \includegraphics[width=0.7\linewidth]{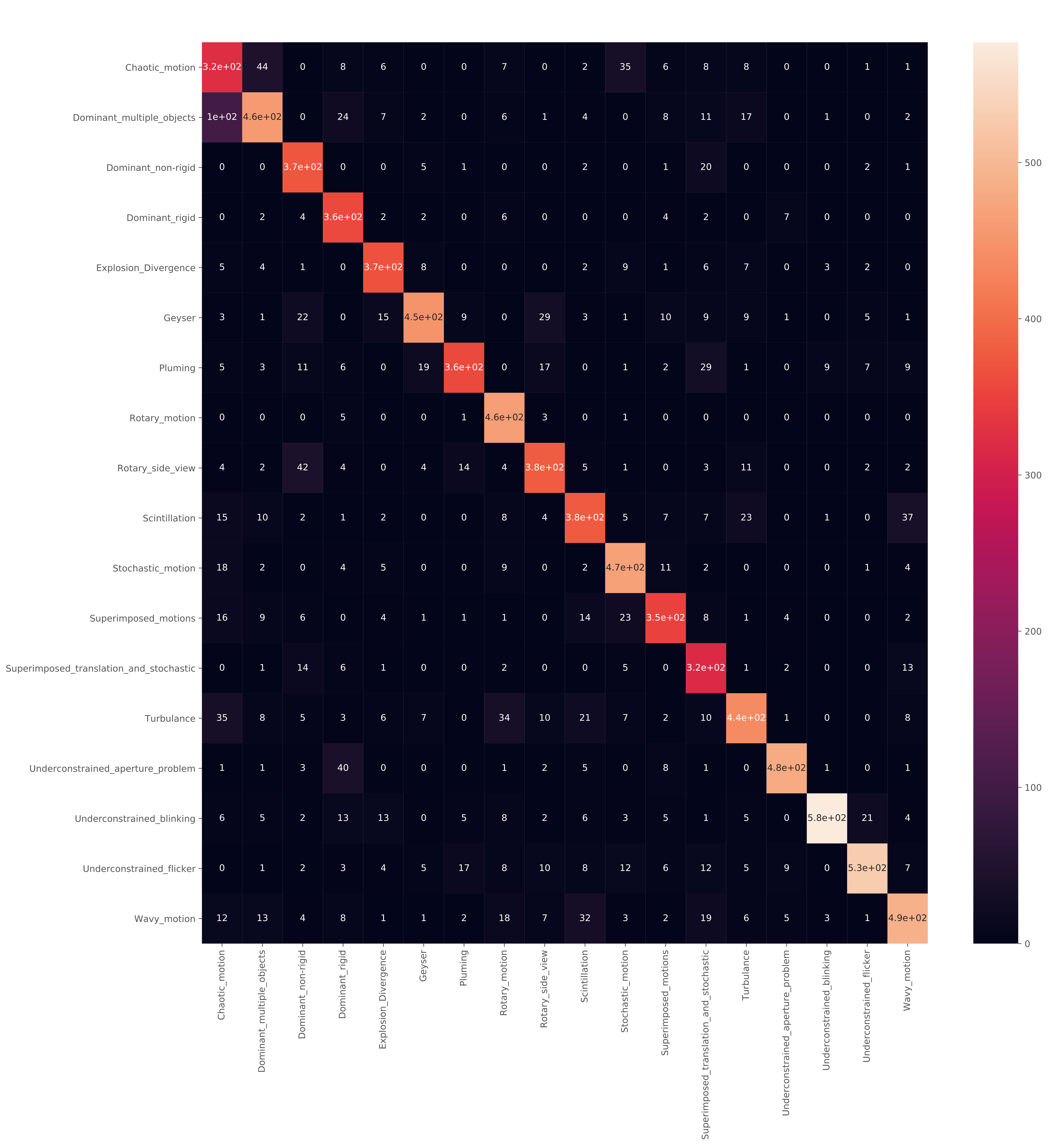}
    \caption{Confusion matrix on the test set of DTDB~\cite{dynamic_dataset} computed from our DTFVD backbone model.}
    \label{fig:dtfvd_confusion_mat}
\end{figure*}

\begin{figure*}
    \centering
    \includegraphics[width=0.6\linewidth]{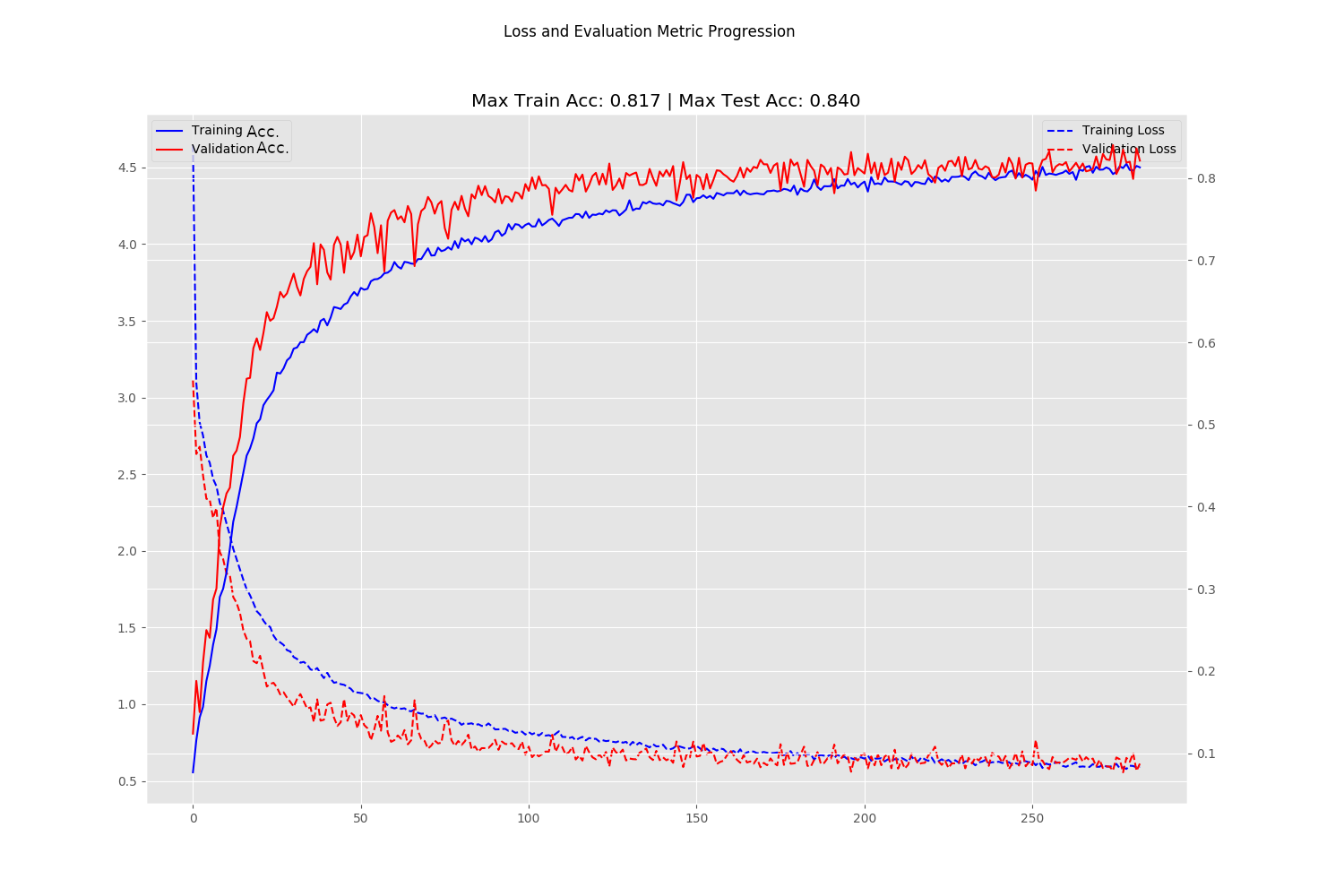}
    \caption{Training and validation loss while optimizing our DTFVD backbone network on a sequence length of $32$. Similar accuracy on both dataset splits indicate a well-generalizing model.}
    \label{fig:dtfvd_loss}
\end{figure*}

In Sec.~4.3 of our main paper, we introduced a dedicated FVD metric for the domain of dynamics textures, the Dynamic Texture Fréchet Video Distance (DTFVD). To this end, we trained a network on DTDB~\cite{dynamic_dataset} for the task of dynamic texture classification. The motivation behind introducing DTFVD is to provide an additional metric which is sensitive to the types of appearances and dynamics encapsulated by dynamic textures, rather than human action-related motions, as captured by FVD.
For the DTFVD network, we use the same architecture as used for the FVD model, i.e., an I3D network \cite{inception}. At convergence (cf. Fig.~\ref{fig:dtfvd_loss}), the DTFVD model achieved $81.7\%$ training accuracy, while achieving $84.0\%$ test accuracy, thus indicating that the model yields well generalizing features capturing the appearance and dynamics in DTDB. A similar conclusion can be drawn by looking at the confusion matrix in Fig.~\ref{fig:dtfvd_confusion_mat} computed for the test set of DTDB, which shows a dominant diagonal structure. Note, we used dropout with a probability of $p=0.5$ to avoid overfitting, which explains why the classification performance is higher on the test set than on the training set. To evaluate sequences with lengths of 16 as well as 32 we train two separate networks.

